\documentclass[10pt,twocolumn,letterpaper]{article}

\usepackage{cvpr}
\usepackage{times}
\usepackage{epsfig}
\usepackage{graphicx}
\usepackage{amsmath}
\usepackage{amssymb}

\usepackage{textcomp}
\usepackage{mathtools}
\usepackage{gensymb}
\usepackage{color}
\usepackage{adjustbox}

\usepackage[numbers,sort,compress]{natbib}
\usepackage{support-caption}
\usepackage{subcaption}

\usepackage{float}

\usepackage{mathtools}
\usepackage{siunitx}
\usepackage{booktabs}
\usepackage{adjustbox}
\usepackage{hyphenat}

\usepackage[final]{microtype}
\usepackage{url}
\usepackage[capitalize]{cleveref}

\DeclarePairedDelimiter\func{(}{)}
\DeclarePairedDelimiter\set{\{}{\}}

\DeclareMathOperator*{\argmin}{\arg\!\min}
\DeclareMathOperator*{\atantwo}{atan2}

\newcommand{\mflate}{Inflating with Maps}

\newcommand{\R}[1]{\mathbb{R}^{#1}}
\newcommand{\med}{s^{k}}

\DeclarePairedDelimiter\abs{\lvert}{\rvert}
\DeclarePairedDelimiter\norm{\lVert}{\rVert}
\DeclareMathOperator\sgn{sgn}

\makeatletter
\let\oldabs\abs{}
\def\abs{\@ifstar{\oldabs}{\oldabs*}}
\let\oldnorm\norm{}
\def\norm{\@ifstar{\oldnorm}{\oldnorm*}}
\let\oldset\set{}
\def\set{\@ifstar{\oldset}{\oldset*}}
\let\oldfunc\func{}
\def\func{\@ifstar{\oldfunc}{\oldfunc*}}
\makeatother

\cvprfinalcopy %

\def\cvprPaperID{****} %

\begin{document}

\title{3D for Free: Crossmodal Transfer Learning using HD Maps}

\author{Benjamin Wilson \({}^{1,2}\)
\qquad
Zsolt Kira \({}^{2}\)
\qquad
James Hays \({}^{1,2}\) \\\\
\({}^{1}\) Argo AI, \({}^{2}\) Georgia Tech \\
}

\twocolumn[\maketitle\vspace{-2em}    \centering
    \captionsetup{type=figure}
    \includegraphics[width=\linewidth]{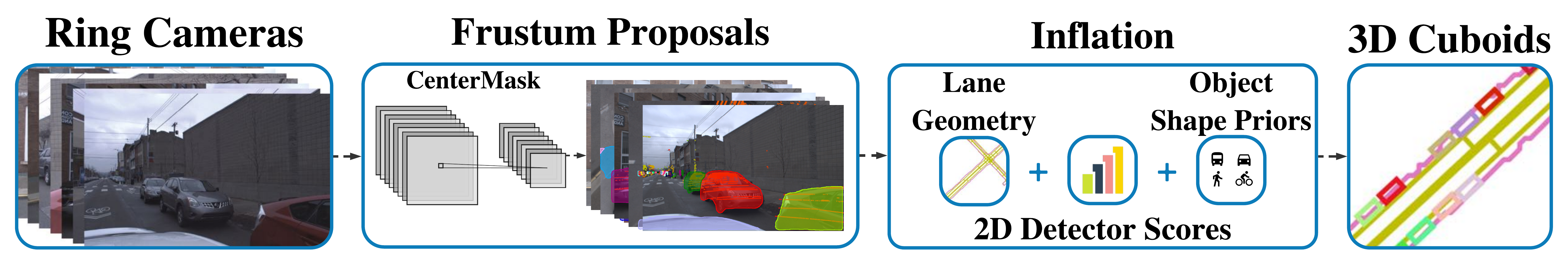}
    \captionof{figure}{\textbf{3D inflation algorithm}. Our method uses synchronized LiDAR and images as input. We project 2D instance segmentation masks onto 3D LiDAR points to seed initial 3D object candidates. The frustum proposals, i.e., instance masks which have been backprojected into 3D, are refined by object shape priors and their orientation is determined by the direction of the closest lane segment obtained from HD maps. Lastly, non-maximum suppression is used to combine redundant cuboids.}
    \label{fig:inflated_architecture}
\bigbreak]

\begin{abstract}
    3D object detection is a core perceptual challenge for robotics and autonomous driving. However, the class-taxonomies in modern autonomous driving datasets are significantly smaller than many influential 2D detection datasets. In this work, we address the ``long-tail'' problem by leveraging both the large class-taxonomies of modern 2D datasets and the robustness of state-of-the-art 2D detection methods. We proceed to mine a large, unlabeled dataset of images and LiDAR, and estimate 3D object bounding cuboids, seeded from an off-the-shelf 2D instance segmentation model. Critically, we constrain this ill-posed 2D-to-3D mapping by using high-definition maps and object size priors. The result of the mining process is 3D cuboids with varying confidence. This mining process is itself a 3D object detector, although not especially accurate when evaluated as such. However, we then train a 3D object detection model on these cuboids, consistent with other recent observations in the deep learning literature, we find that the resulting model is fairly robust to the noisy supervision that our mining process provides. We mine a collection of 1151 unlabeled, multimodal driving logs from an autonomous vehicle and use the discovered objects to train a LiDAR-based object detector. We show that detector performance increases as we mine more unlabeled data. With our full, unlabeled dataset, our method performs competitively with fully supervised methods, even exceeding the performance for certain object categories, without any human 3D annotations. 
\end{abstract}

\section{Introduction}

Deep learning has achieved incredible success across a number of 2D and 3D scene understanding tasks. This progress has been largely due to supervised learning methods that train on large amounts of \emph{labeled} data. In this paper, we are focused on the task of 3D object cuboid detection from LiDAR point clouds. While real-world, 3D-labeled datasets for this task exist, such as the seminal KITTI dataset \citep{geigerAreWeReady2012} and others recently released~\cite{caesarNuScenesMultimodalDataset2020,changArgoverse3DTracking2019,sunScalabilityPerceptionAutonomous2019,lyft2019,ramanishkaDrivingSceneUnderstanding2018}, they require a considerable amount of human annotation effort. New larger 3D-datasets such as Waymo~\cite{sunScalabilityPerceptionAutonomous2019} and nuScenes~\cite{caesarNuScenesMultimodalDataset2020} still have less than 10\% of unique object instances as the 2D COCO dataset~\cite{linMicrosoftCOCOCommon2014}.

In this work, we show that it might be possible to train competitive 3D object detection models without the need for 3D human annotation. We do so by relying on the maturity of 2D instance segmentation methods and another trend in self-driving datasets --- \emph{high-definition (HD) maps}. Recent datasets such as Argoverse~\cite{changArgoverse3DTracking2019}, nuScenes~\cite{caesarNuScenesMultimodalDataset2020}, and Lyft~\cite{lyft2019} contain HD maps that encode lane geometry. HD maps are needed for ``Level 4'' autonomy operations, so they are widely available in cities where fleets operate. It is easy to capture unlabeled driving sensor data localized to HD maps during normal fleet operations.

In this paper, we show that with accurate 2D instance segmentation and HD maps, we can \emph{self-annotate} unlabeled driving logs. Throughout this work, we refer to the term \emph{inflate} as the backprojection of 2D sensor data to 3D with crossmodal data.
We inflate 2D instance segmentations in image space into 3D cuboids in LiDAR space using the lane geometry present in Argoverse's HD maps to constrain otherwise ambiguous orientations.
We can inflate large amounts of data because it requires no human 3D supervision and instead relies on HD maps.
The self-annotated, inflated cuboids can then be used to train a 3D object detector, which is surprisingly accurate --- approaching traditionally supervised methods for popular classes, and \emph{exceeding} traditionally supervised methods for rare classes. \\

\noindent\textbf{Relationship to unsupervised, self-supervised, and transfer learning.}
Many recent works \cite{caronUnsupervisedPreTrainingImage2019,wulffTemporalInterpolationUnsupervised2018} demonstrate that large, unlabeled datasets can be valuable for training or pretraining deep networks. Our method is similar in that it relies on a large, unlabeled dataset. However, our method does not use a ``pretext task'' such as colorization~\cite{zhangColorfulImageColorization2016} or image context prediction~\cite{doerschUnsupervisedVisualRepresentation2015} that can be naturally supervised from the source data. Instead, we ``inflate'' 2D instance segmentations from a state of the art, supervised detection method \cite{leeCenterMaskRealTimeAnchorFree2020} during a \emph{mining} phase, and then use the resulting 3D cuboids as supervision for a traditional, LiDAR-based 3D detector. Thus, our method is better thought of as a crossmodality \emph{transfer learning} method that relies on (1) the relative maturity of 2D instance segmentation and corresponding datasets compared to 3D detection and (2) the ease, in 3D, of using HD-map constraints and data augmentation to overcome the ambiguity and low recall of the inflation process. 

Our contributions in this work are as follows:
\begin{itemize}
    \item We demonstrate that 2D instance segmentation models, along with HD maps and object shape priors, can be used to determine the 3D bounding box of objects surprisingly well.
    \item We show that our mining process, even though it is imperfect and noisy, can be used to train 3D object detection models that are more accurate than the \emph{mining} process itself in terms of mAP.
    \item We achieve state-of-the-art 3D detection accuracy on a \emph{modern} dataset, Argoverse --- outperforming supervised methods --- by training on cuboids inflated from our unlabeled dataset.
\end{itemize}

\section{Related Work}

\paragraph{2D weak and self-supervision.}
Several influential works use a free, noisy, or weak signal to generate large amounts of training data. Researchers have proposed a variety of approaches to weakly supervised 2D image classification, object detection, and semantic segmentation.
For 2D object detection, \citet{chenWeblySupervisedLearning2015} explored first learning a relationship graph by querying an image search engine, and then fine-tuning a network with increasingly realistic examples from another source. Like our work, they benefit from deep networks' robustness to label noise~\cite{rolnickDeepLearningRobust2018} --- noise from imperfect search engine results in their case, and noise from an imperfect mining process in our case.  \citet{sunRevisitingUnreasonableEffectiveness2017} explored the relationship between vision-based deep learning and exponentially increasing dataset size with noisy labels. \citet{xuLearningSegmentVarious2015} incorporated image tags, object detections, and partial labels to produce a semantic segmentation model.
Another line of work uses self-supervision, namely, surrogate tasks for which labels are free, such as estimating rotational transformations applied to the images \cite{gidarisUnsupervisedRepresentationLearning2018}, to learn effective downstream features without explicit human annotation \cite{zhangSplitBrainAutoencodersUnsupervised2017,renCrossDomainSelfSupervisedMultiTask2018}.

\paragraph{3D weak and self-supervision.}
Compared to 2D, fewer methods have examined exploiting weak supervision or self-supervision in 3D. \citet{chenBeatMTurkersAutomatic2014} uses weak 3D annotations to predict high-quality 2D segmentation masks --- the opposite direction that we examine. \citet{stutzLearning3DShape2018} learns a ``weakly supervised'' 3D shape completion model from synthetic data. A number of methods have been proposed to simultaneously leverage \textit{multiple modalities of data}.
This can be done to improve detection performance even under the fully supervised setting \cite{xuPointFusionDeepSensor2018,choMultisensorFusionSystem2014,lahoud2DDriven3DObject2017,qiFrustumPointNets3D2018}. 
The preceding works use a range of strong priors, ranging from CAD models to synthetic data, to reduce the need for labels. In contrast, in this paper, we propose a method to mine large amounts of unlabeled data and estimate the shape and heading of 3D objects using only 2D instance segmentation, HD maps, and simple object size priors. Our method relies on the assumption that objects on the road are likely following the direction of their respective lane.

\paragraph{2D to 3D supervision.}
State-of-the-art 2D detection models~\cite{renFasterRCNNRealTime2015,heMaskRCNN2017} perform extremely well in generalizing to novel environments, making them an appealing tool to utilize even for 3D scene understanding.
On the KITTI leaderboard, two of the top models for car detection fuse point clouds with proposals from a 2D instance detection model \cite{duGeneralPipeline3D2018,qiFrustumPointNets3D2018}.
While some fusion methods perform well, they are still trained on human annotated 3D data.
There are a number of \emph{concurrent} works that also ``inflate'' their 2D detections with crossmodal data. \citet{qinWeaklySupervised3D2020} proposes a density-based 3D proposal module to generate object proposals and additionally use geometric priors. \citet{mengWeaklySupervised3D2020} generates cylindrical proposals under weak supervision from human annotators, subsequently refining and generating the final cuboids.
\citet{zakharovAutolabeling3DObjects2020} proposes a fully differentiable renderer using signed distance fields in conjunction with normalized object coordinates for projective and geometric alignment. In contrast, we leverage the semantic maps, lane geometry, and ground height in HD maps to localize objects in 3D space \emph{without} human annotation or synthetic 3D models. Additionally, we propose a simple modification to anchor-based 3D-object detection models to selectively penalize the network on ``confident'' detections, mined from a large set of \emph{unlabeled} multimodal driving data.
In this paper, we also take a fusion approach with additional map information to automatically mine large amounts of 3D object detection annotations.

\paragraph{Maps for 3D object detection.}
Numerous works in 3D object detection have solely relied on ground plane estimation as a prior to determine where objects should exist \cite{chen3DObjectProposals2018,beltranBirdNet3DObject2018,chenMonocular3DObject2016}.
\citet{yangHDNETExploitingHD2018} utilized both ground and road priors, as well as online map estimation to enhance detection performance. More recently, datasets have included HD maps, which include rasterized driveable area and vector lane segments \cite{changArgoverse3DTracking2019}, but we are not aware of works that leverage these map attributes for 3D object detection.

\paragraph{Autonomous driving datasets.}
With the growing interest in autonomous vehicles, numerous datasets have been released in the last two years \cite{caesarNuScenesMultimodalDataset2020,changArgoverse3DTracking2019,sunScalabilityPerceptionAutonomous2019,lyft2019,ramanishkaDrivingSceneUnderstanding2018}. One consistent trend within the community is the increasing number of available sensor modalities provided within their respective datasets. nuScenes~\cite{caesarNuScenesMultimodalDataset2020} initially provided a detection dataset: 1000 logs with camera, LiDAR, and RADAR modalities. Argoverse~\cite{changArgoverse3DTracking2019} emphasized high-definition maps in their tracking and forecasting datasets, providing rasterized ground height and driveable area, as well as a vector map of centerlines.

\section{Problem Setup}
Autonomous fleets capture large quantities of multimodal data during normal testing. Due to time and budget constraints, it is infeasible to annotate all the captured data, so it goes largely unutilized. With that in mind, our goal is to mine object cuboids without the help of human supervision. Then, we use the mined instances to serve as noisy ``ground truth'' annotations to train a 3D object detection model. Below, we define the 3D object detection task and discuss the multimodal data we will operate on.

\paragraph{LiDAR-based 3D detection.} The promise of robotics-centered applications has sustained a large interest in 3D scene understanding. Understanding \textit{where} objects are in the environment is crucial for downstream tasks, such as tracking, forecasting, and motion planning.
LiDAR sensors are widely used to produce accurate 3D data. A LiDAR sensor samples the surfaces of unoccluded objects and scenery by a time-of-flight measurement, which produces a set of points from the sensor:
\begin{align}
    p_i = (x_i, y_i, z_i), \quad \set{p_i}_{i=1}^{M} \subseteq \R{3},
\end{align}
where modern sensors capture on the order of $M=100,000$ points in each sweep of the scene. In 3D object detection, we endeavor to accurately regress both the size and pose of all objects. This is commonly done by instantiating cuboids, the 3D analog of bounding boxes, which are commonly parameterized as:
\begin{align}
    \begin{bmatrix}
        x, y, z, w, l, h, \theta
    \end{bmatrix}^\top
    \in \R{7},
\end{align}
where \((x, y, z)\) refers to the center of the cuboid, \((w, l, h)\) is the size of the object, and \(\theta\) is the rotation about the vertical axis.

Given a subset of points belonging to one object (e.g., the points in the frustum of a 2D detection), some of these parameters could be roughly estimated \emph{without 3D supervision} by optimizing a minimum perimeter or area bounding cuboid with computational geometry techniques~\cite{toussaint1983solving}. However, the heading  \(\theta\) is particularly difficult to estimate without supervision --- a cuboid and its 180\degree ~rotation will fit a vehicle equally well.
We address this problem by leveraging lane geometry from HD maps.

\paragraph{Multimodal self-driving vehicle data.}
Combining multiple modalities of data, known as sensor fusion, has been widely explored to overcome the shortcomings of each individual sensor.
For example, LiDAR provides accurate depth and occupancy information, but it suffers from sparsity at a distance.
RGB imagery provides a dense grid with detailed texture information; however, lighting sensitivity, lack of size invariance, and the absence of depth information cause complications with object pose and shape estimation.
Map information yields semantic priors for both object location and pose in the world, but detailed maps are not \emph{necessarily} available a priori.
In some sense, sensors can be viewed as either ``online'' or ``offline''. Both LiDAR and RGB act as ``online'' sensors, since they provide \emph{real-time} information of the surrounding scene.
In contrast, HD maps act as an ``offline'' sensor, providing vast amounts of semantic information, albeit at the cost of prior precomputation.
In this work, we explore leveraging both ``online'' and ``offline'' sensors --- LiDAR, RGB, and HD maps --- to mine 3D annotations.
Recent self-driving datasets provide these three types of data~\cite{changArgoverse3DTracking2019,caesarNuScenesMultimodalDataset2020,sunScalabilityPerceptionAutonomous2019}, but we mine 3D cuboids from a larger \emph{unannotated} dataset as outlined in \cref{fig:inflated_architecture}, using Argoverse HD maps. ``HD map'' is a somewhat ambiguous term. We consider providing \emph{lane-level geometry} to be the minimum requirement for a map to be considered ``HD''. Argoverse maps encode lane geometry with a graph of 3D lane centerlines. We will also use Argoverse's ground height map and driveable area map to remove irrelevant LiDAR points.

\section{3D Inflation Method}
In this section, we describe our method for mining objects. Specifically, we discuss producing 3D cuboid-enclosed LiDAR point clouds with synchronized LiDAR and RGB from unlabeled vehicle logs with known ego-vehicle pose on an HD map. 

\paragraph{Seeding the objects of interest --- object frustum proposals.}
We adopt an approach similar to the \textit{frustum proposal generation} method as previously proposed in ``Frustum PointNets''~\cite{qiFrustumPointNets3D2018}.
Frustum proposals are the LiDAR points that fall within the viewing frustum of a particular 2D object proposal. As a preprocessing step, we use HD maps to remove LiDAR points on the ground as defined by the ground height map. We also eliminate LiDAR points outside of the Argoverse Region of Interest --- more than 5 meters from the road --- by using the driveable area map. 
Using Centermask \cite{leeCenterMaskRealTimeAnchorFree2020}, we generate a set of 2D instance proposals in the ring camera images and backproject those masks onto the aligned LiDAR point cloud.
Then, we define a 3D ``seed point'' for the \(k^{\text{th}}\) object proposal by calculating the geometric median, i.e., the point minimizing the sum of distances to its frustum proposal points:
\begin{align}
    \med = \argmin_{y \in \R{3}} \sum_{i=1}^{m_k}\norm*{p_i - y}_2,
\end{align}
where \(m_k\) is the number of points in the \(k^\text{th}\) frustum proposal.
There is no guarantee that the seed point will be located at the ``true'' center of an object. Frequently, the seed point will be close to the visible surface in the object's interior.
However, the seed point anchors our cuboid estimation process and serves as a noisy estimation of the center of an object.

\paragraph{Orientation estimation.} Determining an object's heading without explicit 3D supervision requires careful consideration. As a baseline, we use the ``rotating calipers'' method \cite{toussaint1983solving}, a classical computational geometry algorithm, to compute the minimum area bounding box of an object from a bird's eye view.
With the assumption that the longer axis aligns with the object's heading, we can recover the object's orientation. However, this approach still suffers from \emph{heading} ambiguity, e.g., the above technique arbitrarily chooses one of the two directions along the principal axis as its heading.

\paragraph{Orientation estimation with lane geometry.}
Our goal is to determine a method that estimates the true object heading more accurately than our baseline, the rotating calipers method.
One of our primary assumptions is that the tangent direction of the closest lane serves as a strong prior on the heading of vehicles, buses, and bicycles, thus providing a simple method to estimate the object's orientation \(\theta\) on the ground plane.
First, consider all the points in a scene \(P = \set{p_i}_{i=1}^{M}\).
We project the seed point \(\med\) into the bird's eye view (BEV) to find the tangent direction of the closest lane \(\theta_{L^{k}}\), which we use to determine the object's rotation in the BEV:
\begin{align}
    \theta_{L^{k}} =
    \begin{bmatrix}
    	l_1 \\
        l_2
    \end{bmatrix},
    \quad \theta = \atantwo \func{l_2, l_1}.
\end{align}
\paragraph{LiDAR instance segmentation.}
Noisy instance segmentation masks (i.e., masks including other objects, scenery, or both) lead to inaccurate estimates of the near and far frustum planes, due to incorrect LiDAR instance segmentation.
We mitigate this issue by only considering points which are within a class-conditioned, predefined distance from the object's seed point.
For example, we assume that a vehicle is longer in length than in width.
Given its orientation, we are able to consider points in a region specific to the vehicle class, assuming it extends farther on its first principal axis than its second principal axis. We perform LiDAR instance segmentation from the object frustum proposals as:
\begin{equation}
    \resizebox{.9\columnwidth}{!}{%
    $\begin{aligned}
    S^{k} = \set{
        p_{i} \in P : 
        \func{\abs*{x_i - \med_1} \leq d_{c}^{1}}
        \cdot
        \func{\abs*{y_i - \med_2} \leq d_{c}^{2}}
        \cdot
        \func{z_i \leq d_{c}^{3}}
    }_{i=1}^{M}, \label{eq:far_clip}
    \end{aligned}$%
    }
\end{equation}
where \((d_c^1, d_c^2, d_c^3)\) refers to the distance expanded from the seed point on each principal axis for the \(c^\text{th}\) highest confidence class from the 2D instance segmentation model.
We construct an initial cuboid for the object with the instance segmentation points \(S^{k}\) and its orientation \(\theta\).
Additionally, we update the z center and height to reflect our assumption that the object must touch the ground.

\paragraph{Amodal completion.}
In 3D detection, objects are labeled in two different ways: fitting a ``shrink-wrapped'' cuboid around an object's visible sensor data, or estimating the \emph{unseen} spatial extent of the object based off prior spatial knowledge. The latter method is known as \emph{amodal completion}, a term used in Gestalt theory \cite{leharGESTALTISOMORPHISMQUANTIFICATION1999}.
We adopt the amodal approach due to its ubiquity in modern autonomous driving datasets \cite{caesarNuScenesMultimodalDataset2020,changArgoverse3DTracking2019,sunScalabilityPerceptionAutonomous2019,lyft2019,ramanishkaDrivingSceneUnderstanding2018}.
To this end, we extrapolate the \emph{unseen} spatial extent of the object by extending the object shape prior in each dimension along the object's principal axes.
In contrast to the LiDAR instance segmentation process in \cref{eq:far_clip}, our method stretches cuboids that have localization error due to sensor sparsity or occlusion. We expand each cuboid from the object's radially closest vertex to the ego-vehicle, in the direction of its principal axes. The new cuboid parameters are estimated as:
\begin{equation}
    \resizebox{.9\columnwidth}{!}{%
    $\begin{aligned}
    \hat{d}_{\max}^{k,j} &= 
        \begin{cases}
            \max \Big ( {d_{\max}^{k,j}, d_{\min}^{k,j} + \sgn(d_{\max}^{k,j}) a_{j}^c} \Big ) & \sgn(d_{\max}^{k,j}) > 0 \\
            d_{\max}^{k,j} & \text{otherwise},
        \end{cases} \\
        \hat{d}_{\min}^{k,j} &=
        \begin{cases}
            \max \Big ( {d_{\min}^{k,j}, d_{\max}^{k,j} + \sgn(d_{\min}^{k,j}) a_{j}^c} \Big ) & \sgn(d_{\max}^{k,j}) \leq 0 \\
            d_{\min}^{k,j} & \text{otherwise},
        \end{cases}
    \end{aligned}$%
    }
\end{equation}

for all \(\set{j \in \mathbb{N} : 1 \leq j \leq 3 } \).
We then update the parameter of each object's cuboid as:
\begin{alignat}{3}
    \begin{bmatrix}
        x^{k} \\
        y^{k} \\
        z^{k}
    \end{bmatrix} &=
    \frac{1}{2}
    &\begin{bmatrix}
        \func{\hat{d}_{\max}^{k,1} + \hat{d}_{\min}^{k,1}} \\
        \func{\hat{d}_{\max}^{k,2} + \hat{d}_{\min}^{k,2}} \\
        \func{\hat{d}_{\max}^{k,3} + \hat{d}_{\min}^{k,3}}
    \end{bmatrix}, \\
    \begin{bmatrix}
        w^{k} \\
        l^{k} \\
        h^{k}
    \end{bmatrix} &=
    &\begin{bmatrix} \label{eq:amodal}
        \func{\hat{d}_{\max}^{k,1} - \hat{d}_{\min}^{k,1}} \\
        \func{\hat{d}_{\max}^{k,2} - \hat{d}_{\min}^{k,2}} \\
        \func{\hat{d}_{\max}^{k,3} - \hat{d}_{\min}^{k,3}}
    \end{bmatrix}.
\end{alignat}

\paragraph{Scoring mined objects.}
An associated confidence for each mined object allows us to trade off between precision and recall in the mining process. We simply use the confidence from the 2D detector $c^{k}$ as the confidence of the inflated 3D cuboid $I^{k}$.

\paragraph{Non-maximum suppression.}
Our mining procedure, as described thus far, generates a large number of cuboids, including many of which are overlapping.
To alleviate this issue, we apply non-maximum suppression (NMS), i.e., traditional NMS without the guarantee of axis alignment, from a bird's eye view. 

\begin{figure*}[!htb]
    \centering
    \begin{subfigure}{0.32\textwidth}
        \centering
        \includegraphics[width=\linewidth]{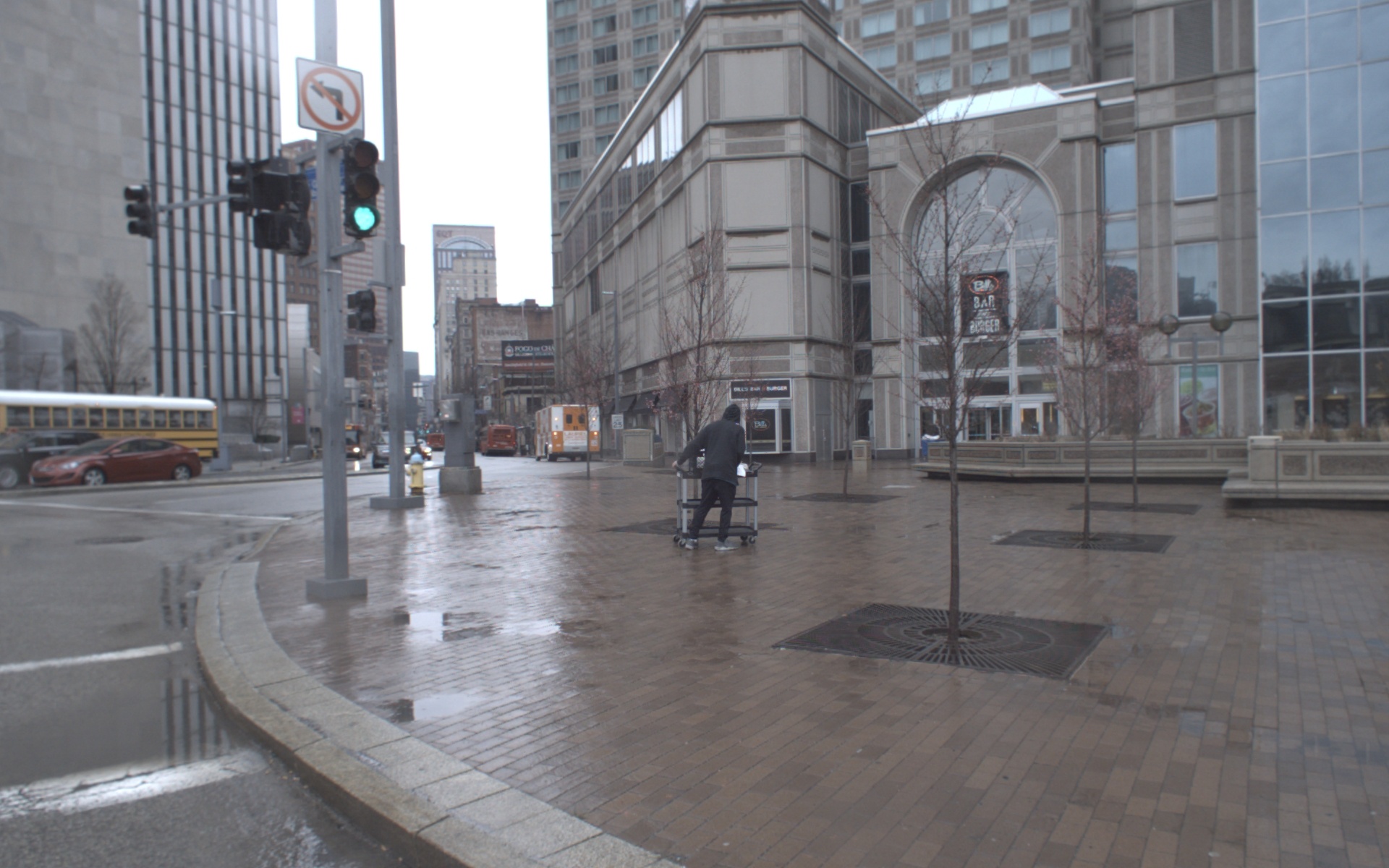}
        \caption{Original camera image}
    \end{subfigure}
    \begin{subfigure}{.32\textwidth}
        \centering
        \includegraphics[width=\linewidth]{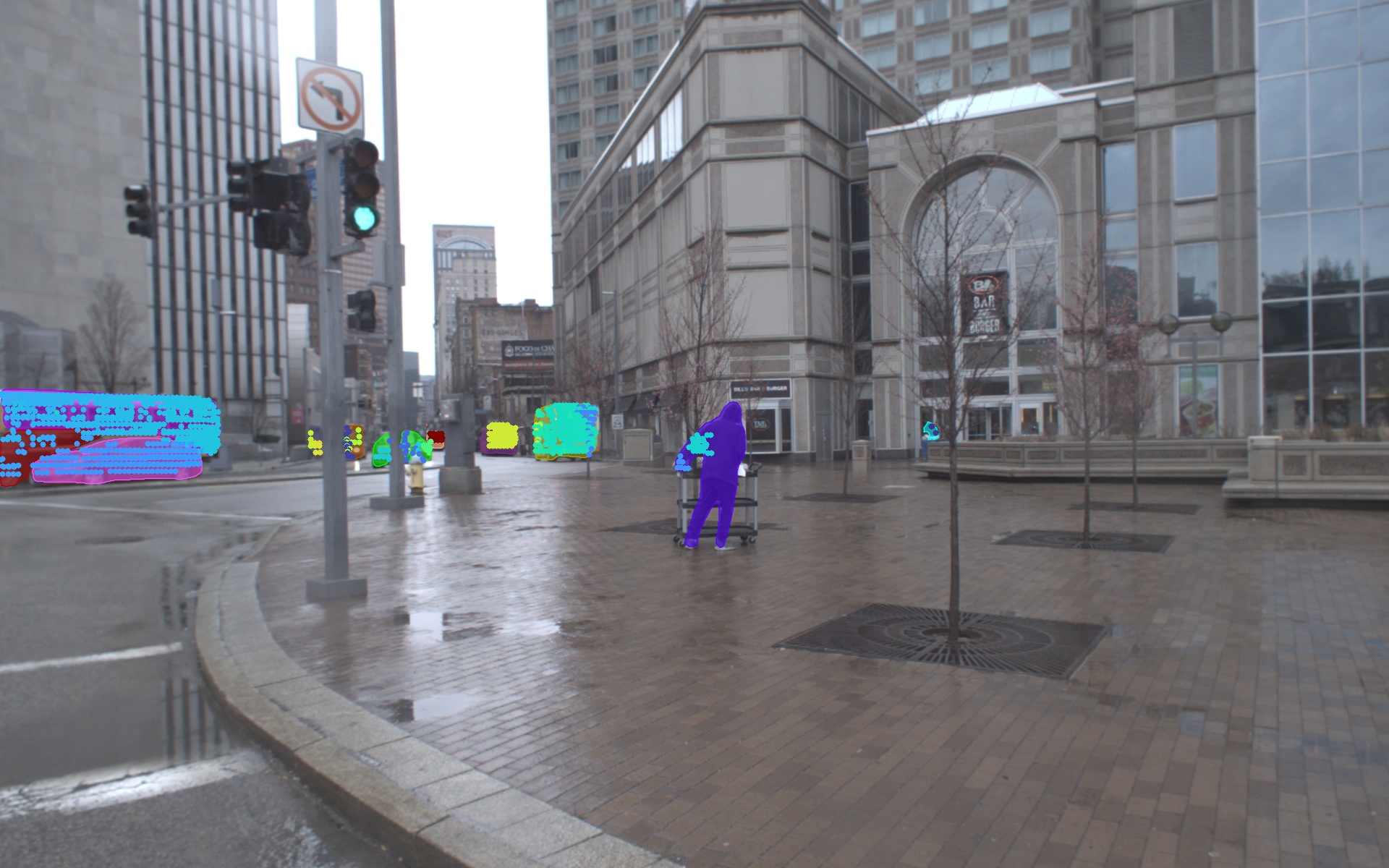}
        \caption{Frustum proposals}
    \end{subfigure}
    
    \begin{subfigure}{.24\textwidth}
        \centering
        \includegraphics[width=\linewidth]{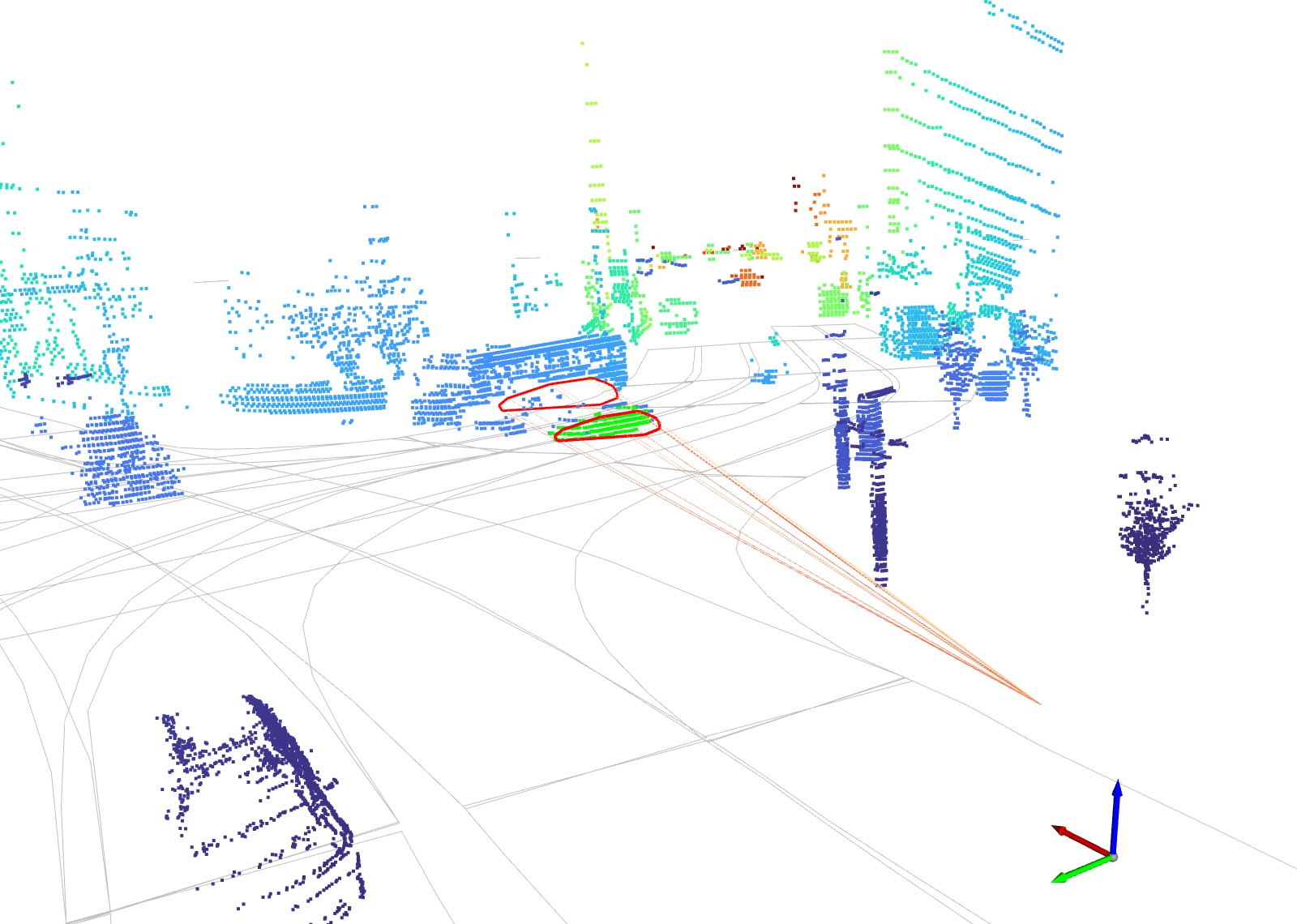}
        \caption{Object frustum proposal}
    \end{subfigure}
    \begin{subfigure}{.24\textwidth}
        \centering
        \includegraphics[width=\linewidth]{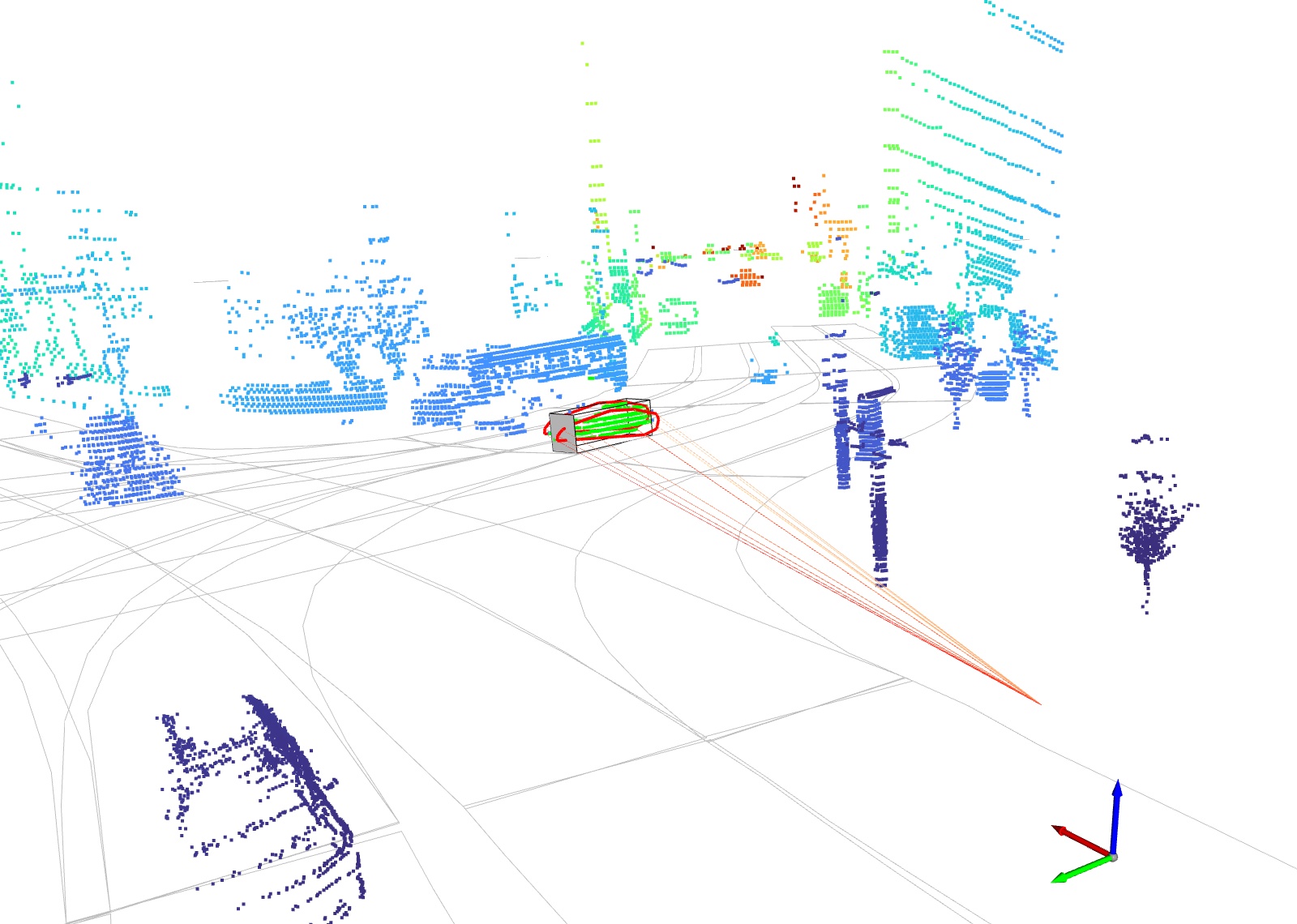}
        \caption{LiDAR instance segmentation}
    \end{subfigure}
    \begin{subfigure}{.24\textwidth}
        \centering
        \includegraphics[width=\linewidth]{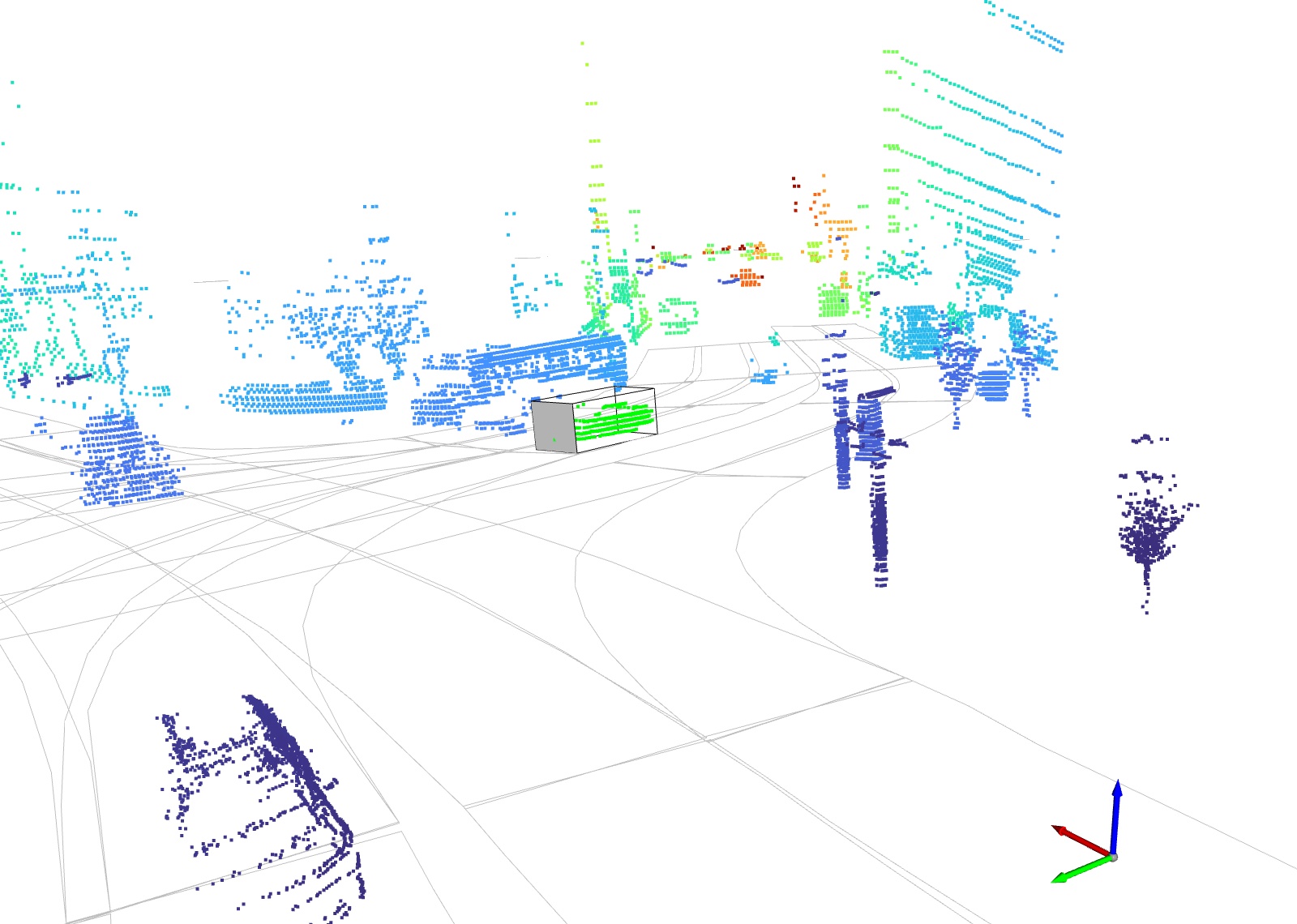}
        \caption{Amodal completion}
    \end{subfigure}
    \begin{subfigure}{.24\textwidth}
        \centering
        \includegraphics[width=\linewidth]{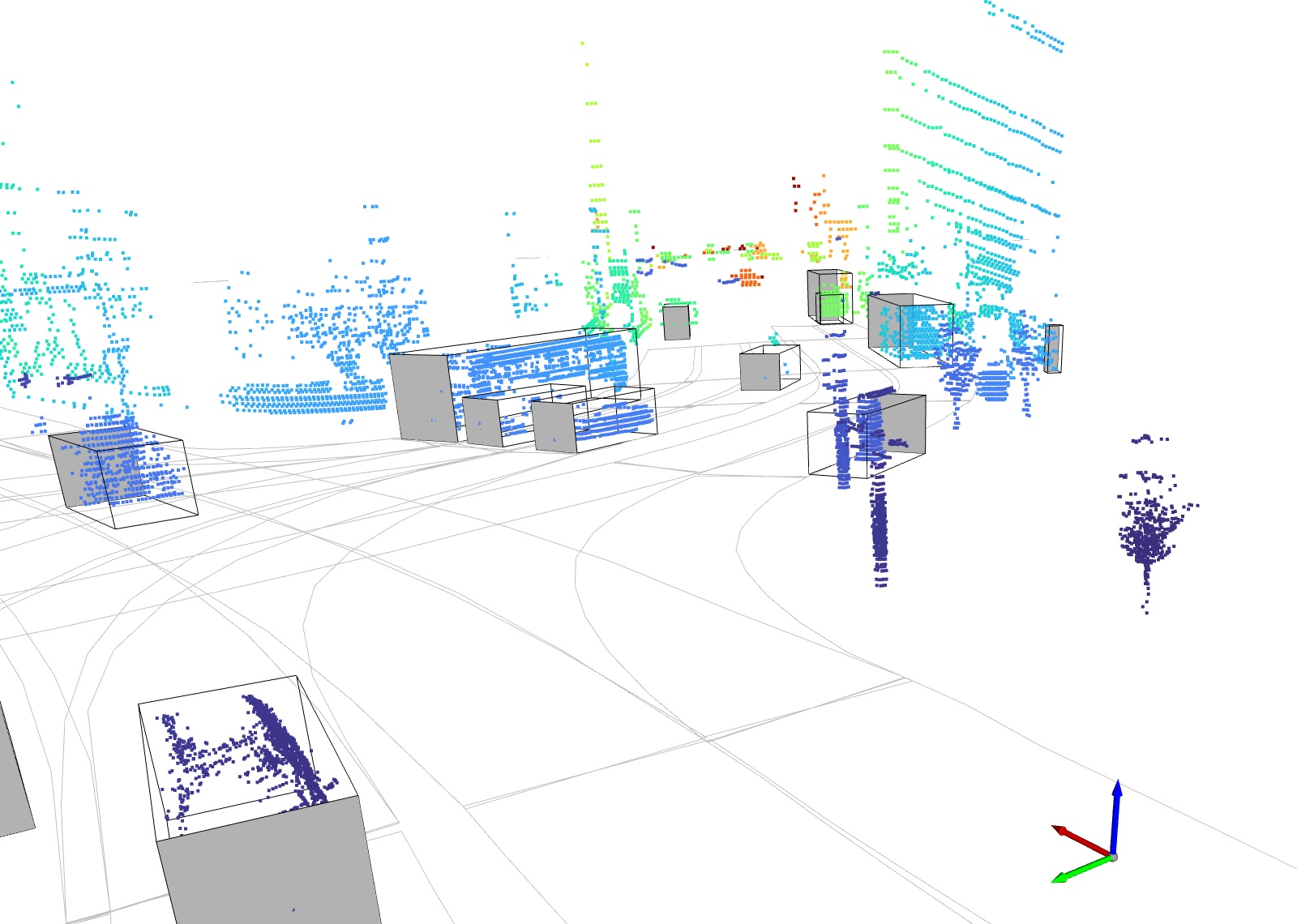}
        \caption{Final cuboids}
    \end{subfigure}
    \caption{\textbf{3D inflation visualization.} We show the stages of our method from two different perspectives: camera imagery and LiDAR sensor data. The coordinate axes are placed at the coordinate frame's origin located at the center of the rear axle of the ego-vehicle. }
    \label{fig:frustum_example}
\end{figure*}

\paragraph{Supervision with inflated data.}
The 2D to 3D inflation process can be thought of as an object detector and we evaluate it as such in the next section.
However, for supervising a 3D detector, we only use the \emph{confidently} mined objects.
Mining \emph{reliable} rather than \emph{diverse} 3D objects is more important because we will use these mined instances to train modern, LiDAR-based 3D detectors that already include sophisticated data augmentation, e.g., cutting, pasting, and rotating object examples inside a scene.
These augmentations are \emph{easier} in 3D than in 2D, where it is difficult to realistically rearrange objects in a scene.

Both of our mining techniques produce a number of false positives.
Clearly, we would not like to \emph{regress} to these cuboids, nor would we like to treat these as \emph{background} points.
To this end, we define ``don't care'' regions, e.g., areas for which assigned anchors don't contribute to the network's loss.
During training, if a cuboid's mined confidence is less than a class-specific threshold, we mark it as ``don't care''.
During training, if an anchor's BEV IoU is greater than the corresponding negative anchor threshold for the respective class, it does not contribute to classification or regression loss.

\section{Experiments}

\paragraph{Datasets.}
To evaluate our method, we use two datasets: the Argoverse 3D tracking dataset and an unlabeled mining dataset that is registered onto Argoverse's HD maps.
While Argoverse is a tracking dataset, we treat it as a detection dataset and do not use any tracking information.
The Argoverse \textit{training} set contains 65 scenes with a total of 13,122 frames, while the \textit{validation} contains 24 scenes with a total of 5015 frames. Our unlabeled \textit{mining} set contains 1151 scenes with a total of 136,248 frames. For these datasets, the LiDAR sensor produces point clouds with an average of 107,000 points at 10 Hz using two 32-beam sensors, while the ring cameras are \((1920 \times 1200)\) at 30 Hz with a 360\degree{} field of view. We evaluate performance on the following classes: vehicle, bus, bicycle, and pedestrian. These classes are chosen due to (a) their overlap with object categories in the pretrained 2D instance segmentation model used and (b) because there are at least 200 Argoverse test set instances within the evaluation range. 

\paragraph{Evaluation metrics.}
For evaluation, we report average precision (AP), average translation error (ATE), average scale error (ASE), and average orientation error (AOE), as defined by the nuScenes object detection challenge~\cite{caesarNuScenesMultimodalDataset2020}.
\begin{figure}[!htb]
    \centering

    \includegraphics[width=.8\linewidth]{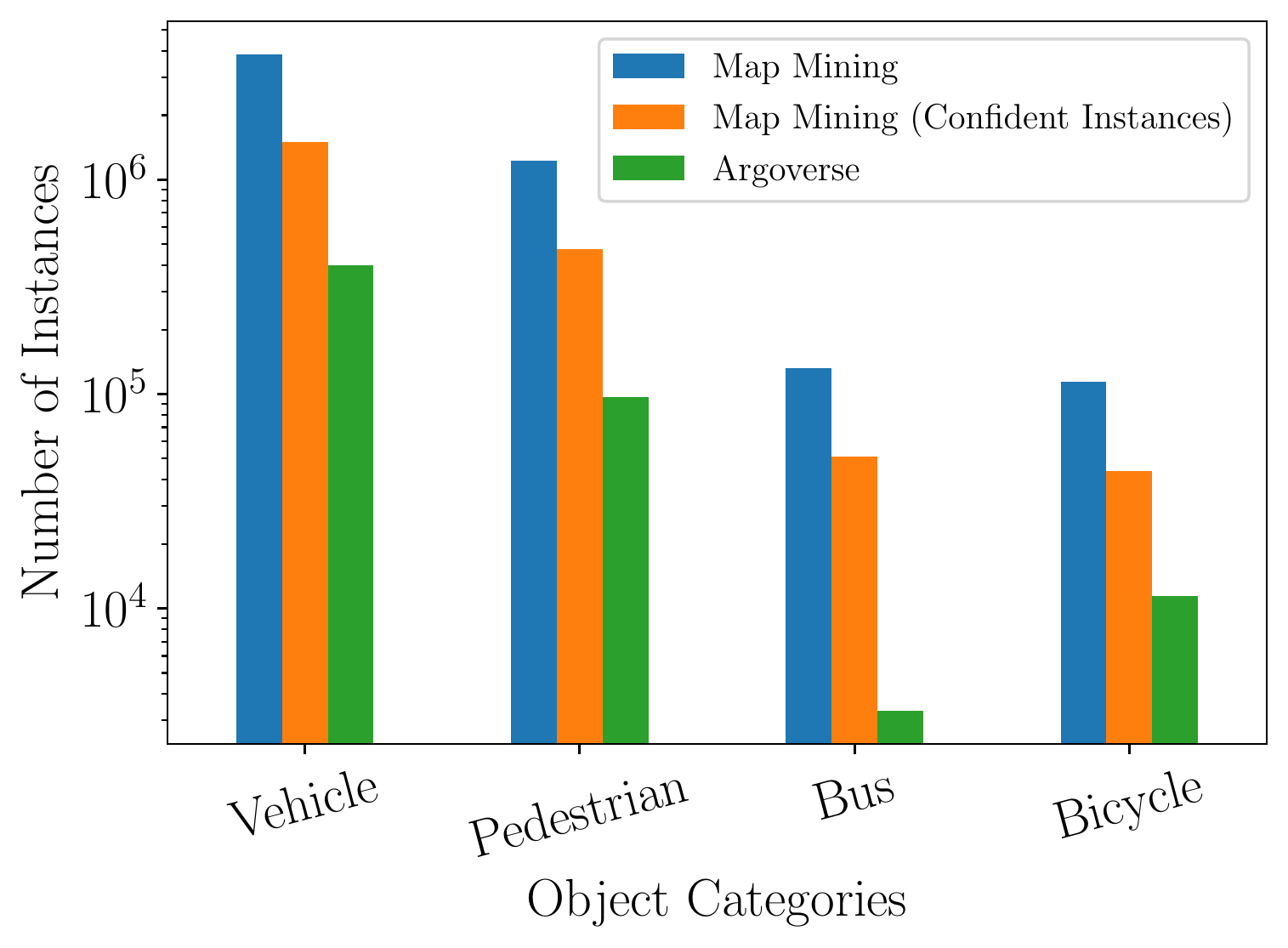}
    \caption{\textbf{Distribution of objects in mining vs. Argoverse dataset.} Rare class detection is a difficult problem within the self-driving community. As shown above, the Argoverse \emph{training} split has an order of magnitude fewer examples of the bus and bicycle categories than the vehicle and pedestrian categories. Our mining process finds rare objects in a much larger, unlabeled mining dataset. After mining, we find 5x to 10x as many object instances, but these mined labels are not as trustworthy as those created through human annotation.}
    \label{fig:dataset_distributions}
\end{figure}

\paragraph{Supervised baseline.}
We compare the performance of our model against a fully supervised baseline.
We use the Class-balanced Grouping and Sampling architecture (CBGS) described in \citet{zhuClassbalancedGroupingSampling2019},
because it is the top performing method on the nuScenes leaderboard with open source code at the time of writing.
CBGS consists of 3 main components: a 3D backbone, a region proposal network (RPN), and a multigroup head. Following the work of \citet{yanSECONDSparselyEmbedded2018}, the 3D backbone consists of both regular and submanifold sparse convolutions. The features are fed into a variant of the RPN in the VoxelNet architecture proposed by \citet{zhouVoxelNetEndtoEndLearning2018}.
Lastly, they propose a multigroup head to learn shared representations for sufficiently similar classes.
CBGS improves performance on all object categories, but especially improves rare object category performance.
We train our baseline on the full Argoverse training set, consisting of 65 human annotated logs.

\paragraph{Evaluating map-based inflation.}
Our 2D-to-3D inflation-based mining process can be evaluated as if it is a 3D object detector. We do this by ``mining'' the Argoverse test set where ground truth human-annotated 3D cuboids are already present. We choose our expansion and object shape prior parameters based on average object size with grid search on the validation split of Argoverse. As Table~\ref{tab:full_comparison} shows, the inflation process achieves a mAP of 37.40 on the Argoverse test set. Note that the orientation error, mAOE, is quite low --- lower than the fully supervised baseline --- which indicates that our map heuristics are worthwhile.

\begin{figure}[!htb]
    \centering

    \begin{subfigure}{.49\columnwidth}
        \centering
        \includegraphics[width=\columnwidth]{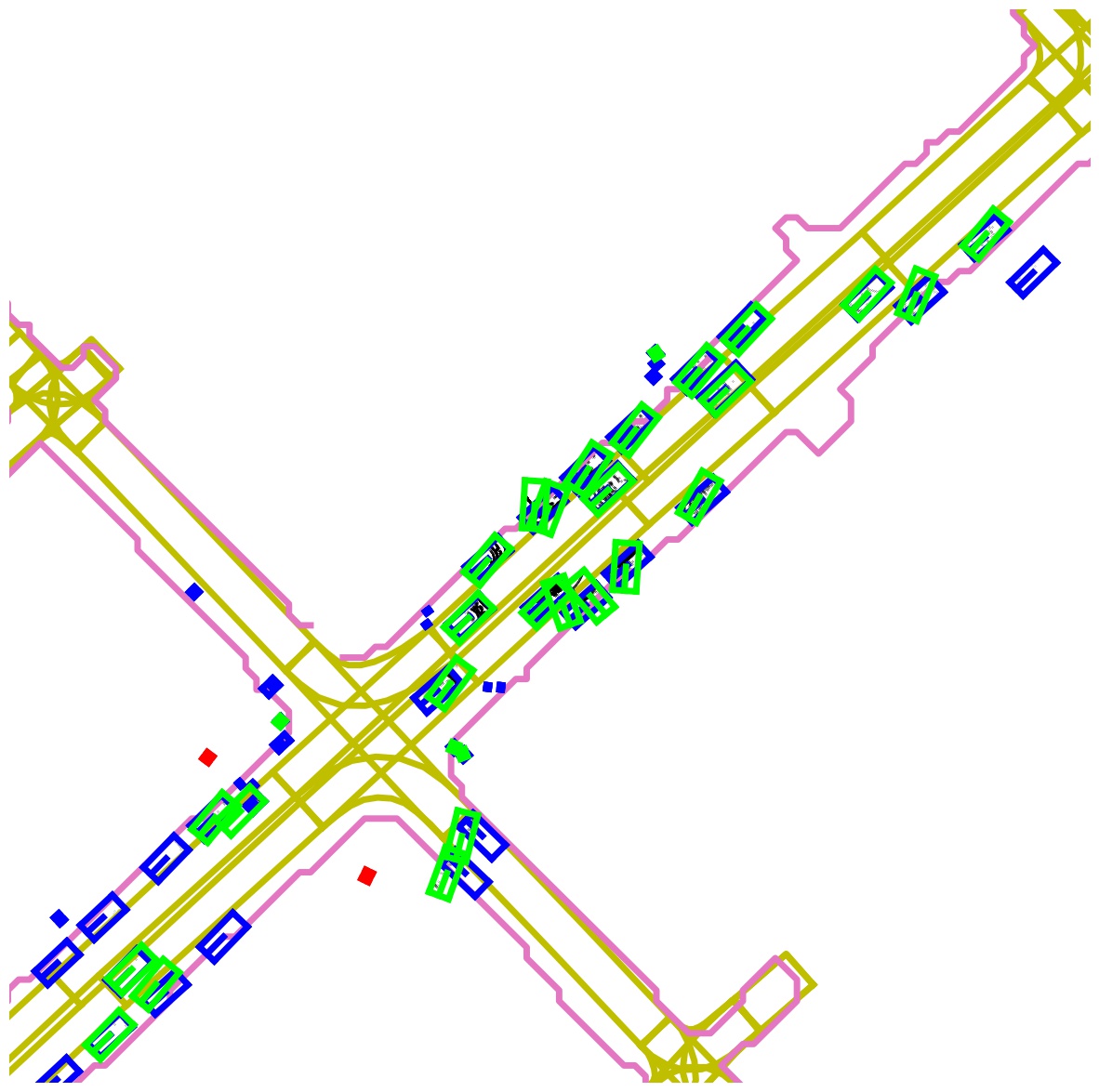}
        \caption{Rotating Calipers}
    \end{subfigure}
    \begin{subfigure}{.49\columnwidth}
        \centering
        \includegraphics[width=\columnwidth]{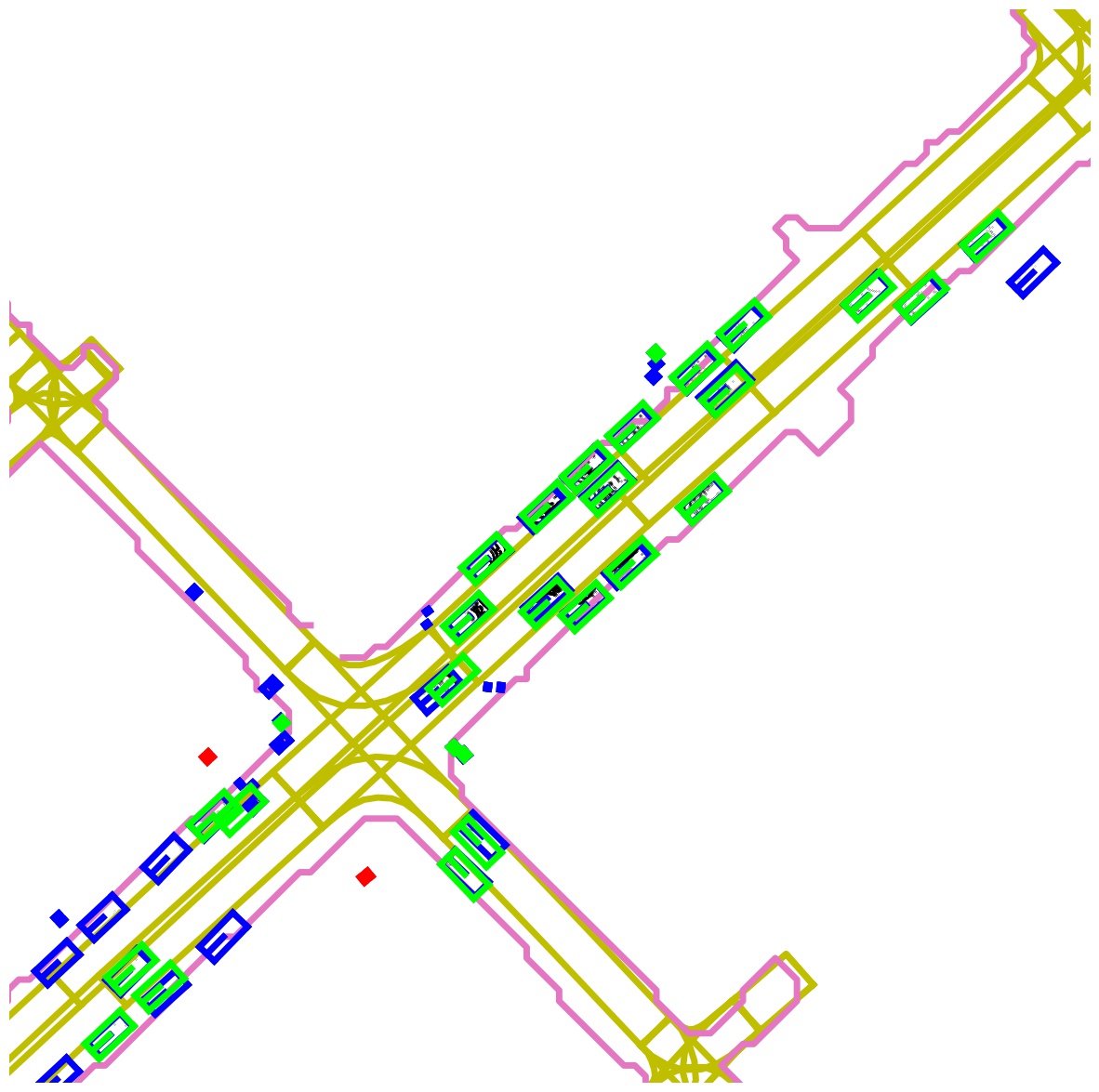}
        \caption{Inflating with Map (ours)}
    \end{subfigure}
    \caption{\textbf{Bird's eye view of inflation methods.} The blue, red, and green boxes denote ground truth, false positive, and true positive, respectively. We use the 2 meter center distance as our criterion for a true positive.}
    \label{fig:miner_visualizations}
\end{figure}

\paragraph{Evaluating geometry-based inflation.}
As a baseline, we also evaluate our inflation method without using the map to orient cuboids. We use the Rotating Calipers method \cite{toussaint1983solving} to find a minimum area bounding box in bird's eye view. Note that our map-based inflation method finds the closest lane to the object to estimate the principal axis \emph{and} the heading of the object. The Rotating Calipers method gives us a principal axis, but not a heading direction, so the heading direction is assigned arbitrarily as one of the two directions along the principal axis. The resulting cuboids are ``amodalized'', but this process is less accurate since the heading estimates have more errors. Table~\ref{tab:full_comparison} shows that using Rotating Calipers instead of lane direction results in a 3 point drop in mAP and increases the orientation error by 50\%. 

\begin{figure}[!htb]
    \centering

    \includegraphics[scale=0.5]{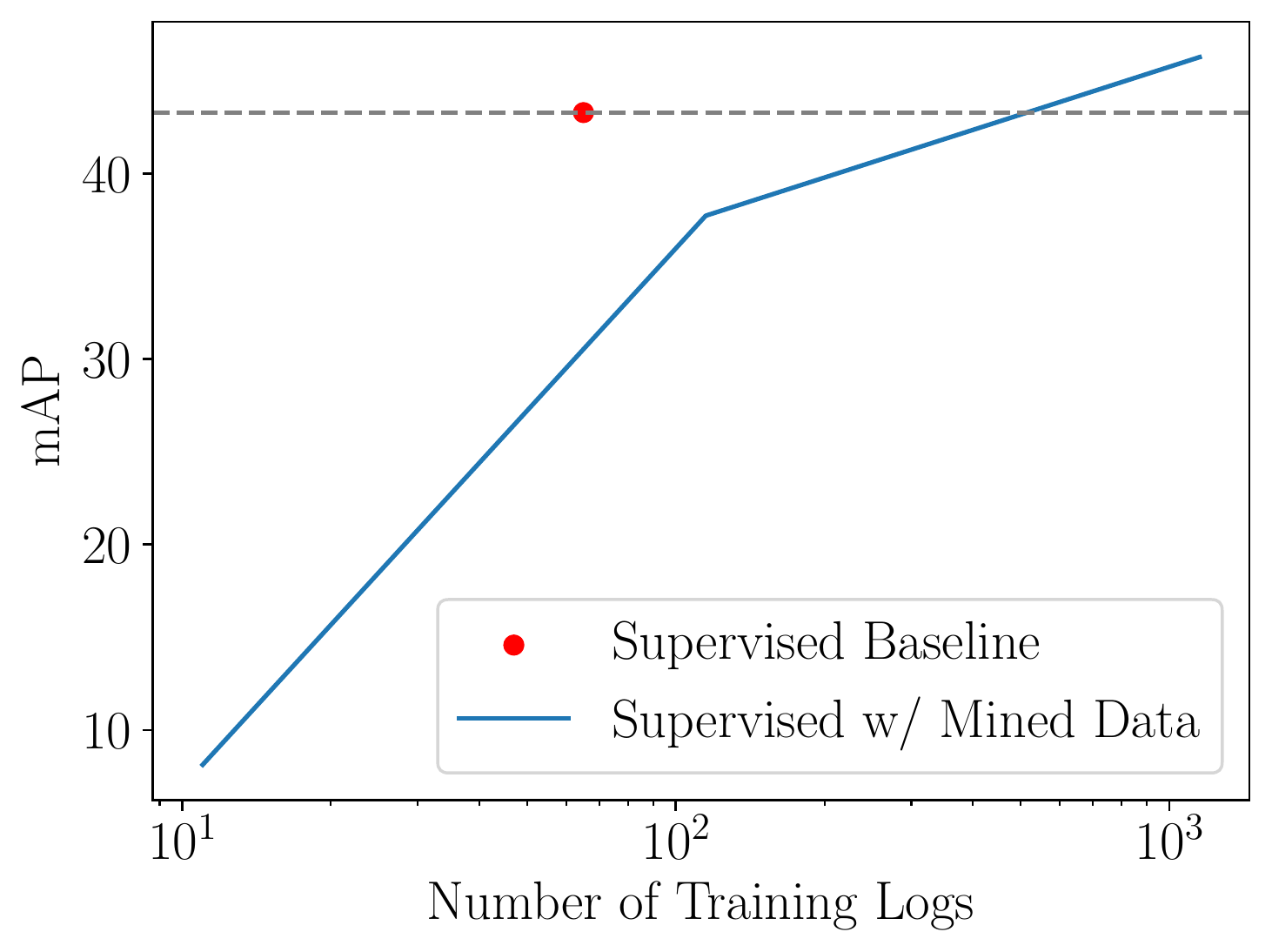}
    \caption{\textbf{mAP vs. Number of Training Logs.} 3D detection performance increases significantly as we train on more mined data. With inflated instances from all 1151 unlabeled scenes, we can supervise a 3D detector that outperforms a baseline method supervised on the human-annotated Argoverse training set of 65 scenes. When mining 10\% (115) of our logs, the resulting detector falls short of the Argoverse baseline (37.74 mAP vs. 43.30 mAP). It is unsurprising that human-annotated 3D labels are more valuable than 3D labels from our mining process, but our mining process requires no human annotation. The results are shown on the Argoverse \emph{test} set.
    }
    \label{fig:subsampling}
\end{figure}

\begin{table*}[!htb]
    \caption{
    \textbf{Detector performance.}
    We evaluate the performance of our method using a subset of the evaluation metrics introduced in the nuScenes dataset \cite{caesarNuScenesMultimodalDataset2020}: mean average precision (mAP), translation error (mATE), scaling error (mASE), and orientation error (mAOE). Our method performs competitively with our supervised baseline, exceeding performance on both mAP and mATE.
    Our method \emph{bootstraps} itself with the mined detections from the \mflate{} method.
    The results are shown on the Argoverse \emph{test} set.
    }
    \label{tab:full_comparison}
    
    \centering
        \begin{tabular}{lcrrrr}
\toprule
                                       Method & Human Annotation &  $\uparrow$ mAP &  $\downarrow$ mATE &  $\downarrow$ mASE &  $\downarrow$ mAOE \\
\midrule
                          Supervised Baseline &       \checkmark &           43.30 &               \textbf{0.30} &               \textbf{0.18} &               0.92 \\
\midrule
             Inflating with Rotating Calipers &                  &           34.31 &               0.54 &               0.33 &               1.35 \\
                           Inflating with Map &                  &           37.40 &               0.41 &               0.31 &               \textbf{0.90} \\
\midrule
 Supervised with Rotating Calipers Mined Data &                  &           39.30 &               0.46 &               0.31 &               1.35 \\
               Supervised with Map Mined Data &                  &           \textbf{46.29} &               0.33 &               0.29 &               1.29 \\
\bottomrule
\end{tabular}
\end{table*}
\begin{table*}[!htb]
    \caption{\textbf{mAP breakdown.} Our method performs significantly better on underrepresented object categories within Argoverse, while still maintaining competitive accuracy on common categories. The results are shown on the Argoverse \emph{test} set.}
    
   	\label{tab:mAP_comparison}
    \centering
    \begin{tabular}{lrrrrr}
\toprule
                                       Method &  Vehicle &    Bus &  Bicycle &  Pedestrian &  $\uparrow$ mAP \\
\midrule
                          Supervised Baseline &    \textbf{86.78} &  19.24 &     1.10 &       \textbf{66.09} & 43.30 \\
\midrule
 Supervised with Rotating Calipers Mined Data &    64.94 &   2.73 &    30.43 &       59.09 & 39.30 \\
                   Supervised with Mined Data &    72.43 &  \textbf{20.24} &    \textbf{31.74} &       60.76 & \textbf{46.29} \\
\bottomrule
\end{tabular}
\end{table*}

\paragraph{Training on inflated data.}
Our goal with the inflation process is to find 3D object instances which are trustworthy enough to train a LiDAR-based 3D detector. As we have seen, the mining process itself does not have an impressive mAP because it inherits any errors from the 2D detector that seeds the process. Still, the most important question for this paper is whether there is enough ``signal'' in potentially noisy mined cuboids to train an effective 3D detector. To this end, we mine a set of 1151 vehicle logs, averaging 15 seconds per log with our map-based Inflation method.
We treat the discovered objects embedded in their scenes as ``pseudo ground truth'' to train a state-of-the-art 3D detection model. 
\cref{fig:dataset_distributions} shows that we mine far more objects than are available in the Argoverse \emph{training} set. This is because we mine 1151 scenes compared to the 65 human-annotated scenes in the Argoverse training set. We use only the ``confident'' subset of the mined cuboids, where the confidence threshold for each class was tuned on the Argoverse \emph{validation} set.

We compare the performance of our 3D detector trained on mined data with the supervised baseline trained on the Argoverse training set in Table \ref{tab:full_comparison}.
Note that our method \emph{does not} use any labeled 3D data, and we evaluate both methods on the Argoverse \emph{test} set.
The detector trained from mined data outperforms the supervised baseline, showing a \emph{2.99\%} absolute increase in mAP and performing competitively in the remaining metrics. 

Table~\ref{tab:mAP_comparison} examines the mAP breakdown per class. The supervised baseline performs better for the vehicle and pedestrian, while the inflation-trained detector performs better for the bus and bicycle. We believe the reason for this is twofold: (i) vehicle and pedestrian are the most common classes, so the supervised baseline is able to train an effective model from the smaller but more trustworthy Argoverse training set and (ii) pedestrians do not orient with traffic lanes as often as the other objects, so our inflation method is less likely to fit accurate cuboids for them.

Additionally, we notice that our method \emph{significantly} outperforms the miner that supplied the data, with an absolute increase in mAP of \emph{8.89\%}. We believe this is (i) because the 3D detector training uses extensive data augmentation to amplify the value of the confident mined objects and (ii) because the errors from the 2D detector that seeded the mining process are often independent of the object appearance in the LiDAR domain. This is similar to the principle behind ``co-training'' across two data modalities~\cite{blumCombiningLabeledUnlabeled1998}. For example, suppose the 2D detector missed every poorly illuminated vehicle. This would decrease the quality of the mined annotations that we use to supervise our LiDAR-based detector. However, LiDAR is illumination invariant, so in the 3D domain these errors would look like randomly dropped labels, and deep learning is very robust to random label noise~\cite{rolnickDeepLearningRobust2018}. The 3D LiDAR-based detector \emph{cannot learn} to reproduce the illumination-based errors of the 2D detector, so it would learn to detect all vehicles, even if poorly illuminated.

Encouragingly, mining more data increases the accuracy of the resulting 3D detector (see \cref{fig:subsampling}). For the same number of scenes, human-annotated 3D data is still more valuable, but it is trivial to collect large amounts of the unlabeled data we use for our mining, which allows us to surpass the supervised baseline.

\begin{table*}[hbt]
    \caption{\textbf{AOE analysis.} The average orientation error largely correlates with our initial assumption -- vehicles and buses will be aligned with the tangent direction of the closest centerline. Bicycle and pedestrian accumulate less error using the Inflating with Rotating Calipers method. We hypothesize that bicycles do not necessarily follow the centerline direction, as we have no guarantee that they are being ridden. The results are shown on the Argoverse \emph{validation} set and the average orientation error is shown in \emph{radians}.}
    
   	\label{tab:AOE_comparison}
    \centering
        \begin{tabular}{lrrrrr}
\toprule
Method &  Vehicle & Bus & Bicycle & Pedestrian & mAOE \\
\midrule
Inflating with Rotating Calipers & 1.11 &  1.27 & \textbf{0.87} & \textbf{1.58} & 1.21 \\
Inflating with Map & \textbf{0.11} &  \textbf{0.02} & 1.41 & 1.60 & \textbf{0.79} \\
\bottomrule
\end{tabular}

\end{table*}

\subsection{Implementation Details}
\label{sec:implementation_details}

\paragraph{2D instance segmentation.}
For 2D instance segmentation, we use an open source implementation of Centermask \cite{leeCenterMaskRealTimeAnchorFree2020}.
We use pretrained COCO weights with the VoVNet-99 backbone \cite{leeEnergyGPUComputationEfficient2019}.
We use all the default parameters, except that we set the instance masks threshold to 0.05. We use the Panoptic Segmentation model due to its higher performance on mask mAP, only using the instance segmentations from its output.
For each LiDAR frame in our unlabeled mining dataset, we detect instance masks in all seven ring camera images at the nearest timestamp. These instance masks will be projected into 3D to seed our cuboid generation process.

\paragraph{COCO to Argoverse class mapping.}

Our method relies on having correspondences between the 2D instance segmentation labels and the 3D objects of interest. We chose the COCO \cite{linMicrosoftCOCOCommon2014} dataset for three reasons: 1. It is ubiquitous in 2D instance segmentation literature. 2. Pretrained weights are available for a number of different models. 3. COCO contains a variety of classes, a few of which overlap with our 3D objects of interest in Argoverse. COCO and Argoverse have 81 and 17 object classes, respectively. For our results, we chose the mapping in Table~\ref{tab:mapping}. COCO contains other classes that could be used, such as mapping all the animals in COCO to the Argoverse animal class, or mapping the COCO motorcycle to the Argoverse motorcycle class, but animal and bicycle are too rare in Argoverse to reliably evaluate such a detector.

\begin{table}[H]
    \caption{\textbf{Class mapping.} Our chosen mapping between the COCO and Argoverse classes. We mapped truck to vehicle due to their large semantic similarity between the two datasets.}
    \label{tab:mapping}
    \centering
    \begin{tabular}{ll}
    \toprule
    COCO       & Argoverse  \\
    \midrule
    Car        & Vehicle    \\
    Truck      & Vehicle    \\
    Bus        & Bus        \\
    Bicycle    & Bicycle    \\
    Person     & Pedestrian \\
    \bottomrule
    \end{tabular}
\end{table}
\paragraph{3D object detection.}
For 3D object detection, we use the author's implementation of CBGS \cite{zhuClassbalancedGroupingSampling2019}. Each object class in our model has a separate detection head. 
Our training configurations are slight modifications of configurations intended for nuScenes. We train for 10 epochs with a learning rate of 0.001 and a batch size of 8, which were chosen through binary search by subsampling the frames in the mined logs and optimizing mAP on the \emph{validation} set of Argoverse.

\paragraph{3D augmentation.}
We augment each scene with a series of augmentations that were originally introduced in a prior 3D object detection work by \citet{yanSECONDSparselyEmbedded2018}.
We follow the suggested augmentation for nuScenes, given by the authors of \citet{zhuClassbalancedGroupingSampling2019}.
We augment each LiDAR frame by pasting objects in other scenes into the current example.
Additionally, we incorporate global rotation noise between \([-\pi/8, \pi/8]\), scaling noise in the range \([0.95, 1.05]\), translation noise of \(\sigma^2 = 0.2\text{m}\) in each axis. Lastly, we shuffle the points of each LiDAR frame and randomly flip the x-axis and y-axis during training.
We choose the learning rate and augmentation scheme for our method via grid search, and we adopt the same augmentation scheme for the supervised baseline for fair comparison.

\paragraph{Compute.}
We train CBGS using 8 NVIDIA V100 GPUs with 32 GB of VRAM each.

\paragraph{Evaluation.}
As previously mentioned, we filter our ground truth detections by adopting nuScenes 2019 CVPR Challenge detection ranges.
The ranges are: \SI[mode=text]{50}{\metre} for vehicles, \SI[mode=text]{50}{\metre} for buses, \SI[mode=text]{40}{\metre} for bicycles, and \SI[mode=text]{40}{\metre} for pedestrians.
These annotations are filtered with respect to the ego-vehicle coordinate frame.

\section{Conclusion}
In this paper, we show that it is possible to mine 3D object cuboids from 2D instance segmentations in unlabeled, multi-modal, self-driving vehicle data. A key enabler of our mining process is HD maps that encode attributes such as ground height and lane geometry. These map elements allow us to reduce the ambiguity in the 2D to 3D ``inflation''. We mine a collection of 1151 unlabeled driving scenes and use the discovered objects to train a detector that outperforms a model trained on the human-annotated Argoverse training set.

{\small
\bibliographystyle{abbrvnat}
\bibliography{egbib}
}

\end{document}